\documentclass[10pt,twocolumn,letterpaper]{article}

\usepackage[utf8]{inputenc}
\usepackage[T1]{fontenc}
\usepackage{times}
\usepackage{amsmath,amssymb}
\usepackage{graphicx}
\usepackage{booktabs}
\usepackage{hyperref}
\usepackage{xcolor}
\usepackage{algorithm}
\usepackage{algorithmic}
\usepackage{caption}
\usepackage{subcaption}
\usepackage{multirow}
\usepackage{enumitem}
\usepackage[margin=2.2cm]{geometry}
\usepackage{microtype}
\usepackage{titlesec}
\usepackage{setspace}
\usepackage{float}
\usepackage{placeins}
\usepackage{stfloats}   

\makeatletter
\renewcommand{\fps@figure}{tb}
\renewcommand{\fps@table}{tb}

\makeatother

\hypersetup{
    colorlinks=true,
    linkcolor=blue,
    citecolor=blue,
    urlcolor=blue,
}

\titleformat{\section}{\Large\bfseries\sffamily}{\thesection.}{0.5em}{}
\titleformat{\subsection}{\large\bfseries\sffamily}{\thesubsection}{0.5em}{}
\titlespacing*{\section}{0pt}{1.5ex plus 0.5ex}{0.8ex plus 0.2ex}
\titlespacing*{\subsection}{0pt}{1.2ex plus 0.4ex}{0.5ex plus 0.1ex}
\newcommand{\method}{\textsc{SpriteToMesh}}
\newcommand{\ie}{\textit{i.e.}}
\newcommand{\eg}{\textit{e.g.}}
\newcommand{\etc}{\textit{etc.}}

\title{
\method{}: Automatic Mesh Generation for 2D Skeletal Animation \\
Using Learned Segmentation and Contour-Aware Vertex Placement
}

\author{
Bastien Gimbert \\
\texttt{firstname.lastname@gmail.com}
}

\date{}

\begin{document}

\maketitle

\begin{abstract}
We present \method{}, a fully automatic pipeline for converting 2D game sprite images into triangle meshes compatible with skeletal animation frameworks such as Spine2D.
Creating animation-ready meshes is traditionally a tedious manual process requiring artists to carefully place vertices along visual boundaries, a task that typically takes 15--60 minutes per sprite.
Our method addresses this through a hybrid learned--algorithmic approach.
A segmentation network (EfficientNet-B0 encoder with U-Net decoder) trained on over 100,000 sprite-mask pairs from 172 games achieves an IoU of 0.87, providing accurate binary masks from arbitrary input images.
From these masks, we extract exterior contour vertices using Douglas-Peucker simplification with adaptive arc subdivision, and interior vertices along visual boundaries detected via bilateral-filtered multi-channel Canny edge detection with contour-following placement.
Delaunay triangulation with mask-based centroid filtering produces the final mesh.
Through controlled experiments, we demonstrate that direct vertex position prediction via neural network heatmap regression is fundamentally not viable for this task: the heatmap decoder consistently fails to converge (loss plateau at 0.061) while the segmentation decoder trains normally under identical conditions.
We attribute this to the inherently artistic nature of vertex placement--the same sprite can be meshed validly in many different ways.
This negative result validates our hybrid design: learned segmentation where ground truth is unambiguous, algorithmic placement where domain heuristics are appropriate.
The complete pipeline processes a sprite in under 3~seconds, representing a speedup of $300\times$--$1200\times$ over manual creation.
We release our trained model to the game development community.
\end{abstract}

\section{Introduction}
\label{sec:intro}

Two-dimensional skeletal animation has become the dominant technique for creating fluid character and object animations in 2D games, mobile applications, and interactive media.
Frameworks such as Spine2D~\cite{spine2d}, DragonBones~\cite{dragonbones}, and Live2D enable artists to rig a single sprite image with a bone hierarchy and deform it through mesh vertices, eliminating the need for frame-by-frame hand-drawn animation.

Central to this pipeline is the creation of a \emph{triangle mesh} that covers the sprite area.
The quality of the mesh directly impacts the smoothness and realism of the resulting animation: vertices must be placed not only along the outer silhouette of the sprite, but also along internal visual boundaries---color transitions, geometric edges, and structural features---so that different parts of the image can be deformed independently.

In current production workflows, mesh creation remains an entirely manual process.
An artist must carefully:
\begin{enumerate}[nosep]
    \item Trace the outer contour of the sprite;
    \item Identify and follow internal visual boundaries (limb edges, clothing seams, facial features);
    \item Place vertices at structurally important positions (corners, joints, curvature maxima);
    \item Sometime triangulate the result while ensuring mesh quality.
\end{enumerate}
This process typically requires 15--60 minutes per sprite, representing a significant bottleneck in game asset production pipelines that may involve hundreds or thousands of individual sprites.

Existing automated approaches are limited.
The auto-meshing tools built into Spine2D and similar software operate only on the alpha channel, producing regular grids or simple convex hulls that ignore the internal visual structure of the sprite entirely.

In this work, we present \method{}, a fully automatic pipeline that converts a 2D sprite image into a triangle mesh suitable for skeletal animation.
Our approach combines deep learning for mask segmentation with classical computer vision algorithms for structure-aware vertex placement.

Vertex positions in animation meshes are fundamentally \emph{artistic decisions} , rather than deterministic functions of pixel values.

\section{Related Work}
\label{sec:related}

\paragraph{2D Skeletal Animation.}
Spine2D~\cite{spine2d} is a widely adopted framework for 2D skeletal animation.
It uses a bone hierarchy to deform triangle meshes overlaid on sprite images, enabling smooth animations from a single artwork.
Other notable frameworks include DragonBones~\cite{dragonbones}, Live2D~\cite{live2d}, and Creature Animation~\cite{creature}.
All of these systems require manual mesh creation, which our work aims to automate.

\paragraph{Image Segmentation.}
Semantic segmentation has been extensively studied in computer vision.
U-Net~\cite{ronneberger2015unet} introduced the encoder-decoder architecture with skip connections that remains the foundation of most segmentation networks.
EfficientNet~\cite{tan2019efficientnet} provided a family of efficient convolutional backbones that achieve strong performance through compound scaling.
The combination of EfficientNet encoders with U-Net-style decoders has proven effective across many segmentation tasks.
In our work, we use this architecture specifically for sprite foreground segmentation, a binary segmentation task with unique challenges due to the extreme diversity of artistic styles encountered in 2D animation assets.

\paragraph{Mesh Generation from Images.}
The problem of generating 2D meshes from images has been studied in several contexts.
Marching squares algorithms~\cite{lorensen1987marching} can extract contours from binary images but produce axis-aligned meshes.
Alpha shapes~\cite{edelsbrunner1983shape} generalize convex hulls but do not consider internal structure.
Constrained Delaunay triangulation~\cite{shewchuk1996triangle} is widely used for quality mesh generation given a set of vertices, but the problem of \emph{where to place} those vertices remains open.
TexturePacker~\cite{texturepacker} and similar sprite-sheet tools provide simple rectangular or convex-hull meshes but do not follow internal visual boundaries.

\paragraph{Contour Detection.}
The Canny edge detector~\cite{canny1986computational} remains one of the most widely used edge detection algorithms.
The bilateral filter~\cite{tomasi1998bilateral} enables edge-preserving smoothing, reducing texture noise while maintaining sharp boundaries.
The Douglas-Peucker algorithm~\cite{douglas1973algorithms} provides an optimal polygonal approximation of curves with user-controlled tolerance.
Our interior vertex placement pipeline combines these classical methods in a novel way: bilateral filtering for noise reduction, multi-channel Canny for robust edge detection, and Douglas-Peucker for structural vertex selection along detected edges.

\paragraph{Vertex Prediction with Neural Networks.}
Several works have attempted to predict keypoint or vertex positions using neural networks, typically through heatmap regression~\cite{newell2016stacked,sun2019deep}.
While successful for anatomical keypoints (human pose, facial landmarks), where positions are constrained by physical structure, our experiments show that this approach fails for animation mesh vertices.
We attribute this to the fundamentally artistic nature of vertex placement in animation meshes: the same image region can be meshed in many valid ways, and the ``ground truth'' reflects individual artist preferences rather than objective image properties.

\section{Method}
\label{sec:method}

\begin{figure}[H]
    \centering
    \includegraphics[width=\linewidth,height=0.16\textheight,keepaspectratio]{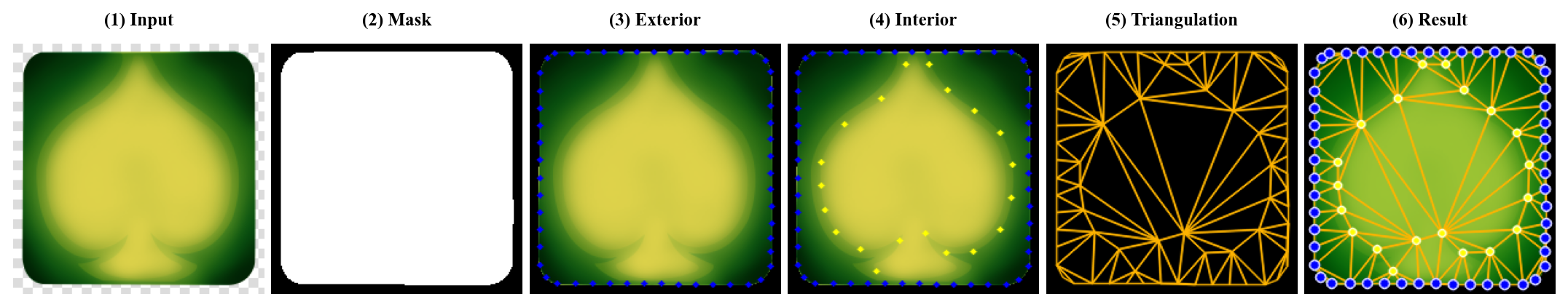}
    \caption{Overview of the \method{} pipeline in six panels. (1)~Input sprite image. (2)~Binary segmentation mask obtained from the alpha channel or our EfficientNet-B0 network. (3)~Exterior contour vertices (blue) placed via Douglas-Peucker simplification with adaptive arc subdivision. (4)~Interior boundary vertices (yellow) detected via multi-channel Canny edge detection, combined with exterior vertices (blue). (5)~Delaunay triangulation wireframe with centroid-based filtering. (6)~Final mesh overlaid on the original sprite.}
    \label{fig:pipeline}
\end{figure}

\method{} takes a single RGB(A) sprite image as input and produces a triangle mesh as output.
The pipeline consists of four sequential stages: (1)~mask acquisition, (2)~exterior contour extraction, (3)~interior boundary detection, and (4)~Delaunay triangulation.
Figure~\ref{fig:pipeline} provides an overview.

\subsection{Mask Acquisition}
\label{sec:mask}

The first stage produces a binary mask $M \in \{0, 1\}^{H \times W}$ separating the sprite foreground from the background.

\paragraph{Alpha Channel Path.}
Many sprite images include an alpha channel that encodes transparency.
When present and non-trivial (\ie, not uniformly opaque), we threshold the alpha channel at $\tau_\alpha = 128$ to produce the binary mask directly:
\begin{equation}
    M(x, y) = \begin{cases} 1 & \text{if } \alpha(x, y) > \tau_\alpha \\ 0 & \text{otherwise} \end{cases}
\end{equation}
This path does not require a learned model and is used whenever possible.

\paragraph{Neural Segmentation Path.}
When no alpha channel is available (\eg, the sprite is composited on a solid or textured background), we use a convolutional neural network to predict the segmentation mask.

Our segmentation network follows a U-Net~\cite{ronneberger2015unet} architecture with an EfficientNet-B0~\cite{tan2019efficientnet} encoder pre-trained on ImageNet~\cite{deng2009imagenet}.
The encoder extracts multi-scale features at five resolution levels through the standard EfficientNet block structure:
\begin{itemize}[nosep]
    \item Blocks 0--1: $\mathbf{s}_1 \in \mathbb{R}^{16 \times 128 \times 128}$
    \item Block 2: $\mathbf{s}_2 \in \mathbb{R}^{24 \times 64 \times 64}$
    \item Block 3: $\mathbf{s}_3 \in \mathbb{R}^{40 \times 32 \times 32}$
    \item Blocks 4--5: $\mathbf{s}_4 \in \mathbb{R}^{112 \times 16 \times 16}$
    \item Blocks 6--8 (bottleneck): $\mathbf{b} \in \mathbb{R}^{1280 \times 8 \times 8}$
\end{itemize}

The decoder consists of four upsampling blocks, each performing bilinear upsampling followed by a $1 \times 1$ convolution, concatenation with the corresponding skip connection $\mathbf{s}_i$, and a double convolution block (Conv$3 \times 3$--BN--ReLU $\times 2$).
The channel progression through the decoder is $1280 \rightarrow 256 \rightarrow 128 \rightarrow 64 \rightarrow 32$.
A final upsampling stage followed by Conv$3 \times 3$--BN--ReLU and a $1 \times 1$ convolution produces the segmentation logits at the original resolution.

The network has approximately 9.3 million parameters total.
The input is a $256 \times 256$ RGB image normalized with ImageNet statistics; the output is resized back to the original image dimensions via bilinear interpolation and thresholded at 0.5.

\paragraph{Training.}
The segmentation loss combines binary cross-entropy and Dice loss:
\begin{equation}
    \mathcal{L}_{\text{seg}} = \mathcal{L}_{\text{BCE}}(\hat{M}, M) + \mathcal{L}_{\text{Dice}}(\hat{M}, M)
\end{equation}
where $\hat{M}$ denotes the predicted mask logits and $M$ the ground truth.
The Dice loss~\cite{milletari2016vnet} is particularly important for handling class imbalance, as sprite foreground typically covers 20--60\% of the image area:
\begin{equation}
    \mathcal{L}_{\text{Dice}} = 1 - \frac{2 \sum_i \hat{p}_i \cdot m_i + \epsilon}{\sum_i \hat{p}_i + \sum_i m_i + \epsilon}
\end{equation}
where $\hat{p}_i = \sigma(\hat{M}_i)$ is the predicted probability at pixel $i$, $m_i$ is the ground truth, and $\epsilon = 1$ is a smoothing constant.

We train using AdamW~\cite{loshchilov2019adamw} with an initial learning rate of $10^{-3}$ for the decoder and $10^{-4}$ for the encoder (lower to preserve pre-trained features), weight decay of $10^{-4}$, and cosine annealing~\cite{loshchilov2017sgdr} over 30 epochs.
The best model is selected based on validation IoU.

\subsection{Why Not Predict Vertices Directly?}
\label{sec:heatmap_failure}

A natural question is whether vertex positions can be predicted directly by the neural network, avoiding the need for hand-crafted vertex placement heuristics.
We investigated this approach thoroughly using heatmap regression.

\paragraph{Architecture.}
We extended the segmentation network with a second decoder head (dual-decoder architecture) dedicated to vertex heatmap prediction.
The heatmap decoder has the same U-Net structure as the segmentation decoder but with an independent set of weights, sharing only the EfficientNet encoder.
The target heatmap is generated by placing 2D Gaussian kernels ($\sigma = 3$ pixels) at each ground truth vertex position, with max-merge to avoid additive saturation at nearby vertices.

\paragraph{Loss.}
The heatmap loss uses weighted MSE to address extreme class imbalance (vertex regions constitute less than 1\% of pixels):
\begin{equation}
    \mathcal{L}_{\text{hm}} = \frac{\sum_i w_i (\hat{h}_i - h_i)^2}{\sum_i w_i}, \quad
    w_i = \begin{cases} w_+ & \text{if } h_i > 0.01 \\ 1 & \text{otherwise} \end{cases}
\end{equation}
where $w_+ = 100$ is the positive pixel weight.
The combined loss is $\mathcal{L} = \mathcal{L}_{\text{seg}} + \lambda \mathcal{L}_{\text{hm}}$ with $\lambda = 50$.

\paragraph{Result: Failure to Converge.}
Despite extensive hyperparameter tuning (varying $\sigma$, $w_+$, $\lambda$, learning rates, and architectures), the heatmap prediction consistently failed to produce meaningful results.
The segmentation head converged normally (IoU reaching 0.87 at epoch 14), while the heatmap loss plateaued at 0.061 without producing localized peaks.
The network learned to predict a near-uniform low activation across the sprite region, failing to identify any specific vertex locations.

\paragraph{Analysis.}
We attribute this failure to the inherently \emph{artistic} nature of vertex placement in animation meshes.
Unlike anatomical keypoints (joints, facial landmarks), whose positions are determined by physical structure, mesh vertex positions reflect individual artist preferences.
The same sprite region can be validly meshed in many different ways, with vertices placed at different densities and positions.
The ground truth vertex positions in our dataset exhibit high variance across different artists and games, making this a fundamentally ambiguous learning target.
This negative result motivated our hybrid approach: leverage deep learning for segmentation (where ground truth is unambiguous) and use classical algorithms for vertex placement (where domain-specific heuristics are appropriate).

\subsection{Exterior Contour Extraction}
\label{sec:contour}

Given the binary mask $M$, we extract the exterior contour and place vertices along it using an adaptive approach that preserves sharp corners while smoothly approximating curved regions.

\paragraph{Step 1: Contour Extraction.}
We apply light Gaussian smoothing ($\sigma = 1$ pixel) to the binary mask to suppress single-pixel noise, then extract the full-resolution contour using OpenCV's \texttt{findContours} with \texttt{CHAIN\_APPROX\_NONE}, which returns every contour pixel.
Multiple connected components are handled by processing each valid contour (area $\geq 100$ pixels) separately, with vertex budget allocated proportionally to perimeter.

\paragraph{Step 2: Douglas-Peucker Simplification.}
The raw contour (typically thousands of pixels) is simplified using the Douglas-Peucker algorithm~\cite{douglas1973algorithms} with tolerance $\epsilon = 0.003 \times P$, where $P$ is the contour perimeter.
This identifies the \emph{structurally important} keypoints: sharp corners, concavities, and inflection points.
The small $\epsilon$ value ensures that all visually significant features are preserved.

\begin{figure}[H]
    \centering
    \includegraphics[width=\columnwidth]{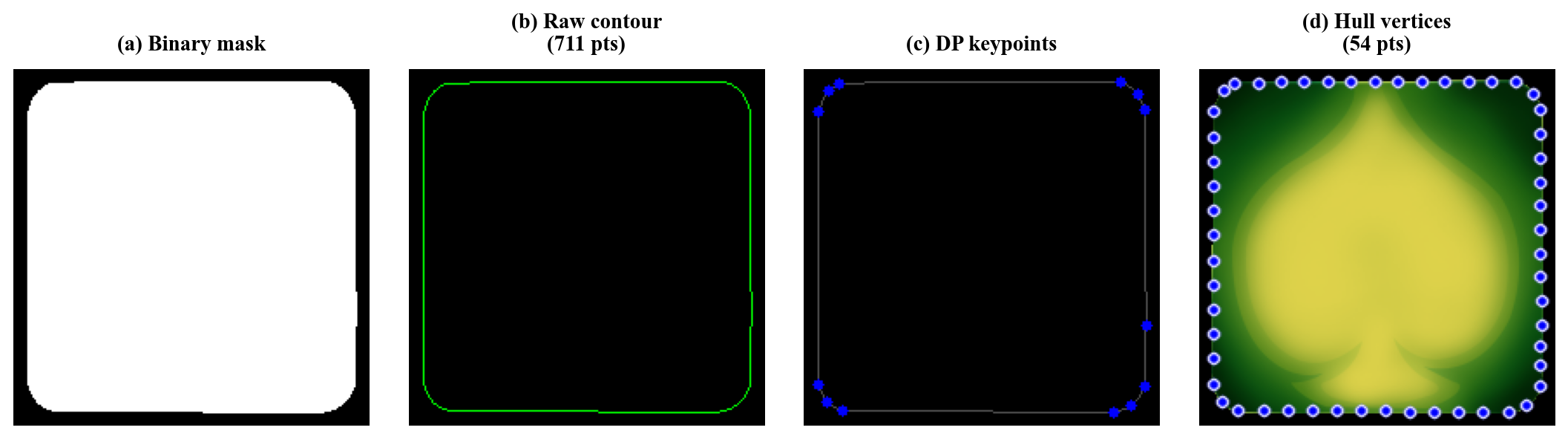}
    \caption{Exterior contour extraction process in four panels. (a)~Binary segmentation mask. (b)~Raw contour extracted with \texttt{findContours}, showing thousands of contour pixels. (c)~Douglas-Peucker simplification ($\epsilon = 0.003 \times P$) identifies structural keypoints at corners and concavities (red dots). (d)~Final hull vertices (red dots with white outlines) after adaptive arc subdivision adds uniformly-spaced vertices between keypoints along the actual contour path.}
    \label{fig:contour_steps}
\end{figure}

\paragraph{Step 3: Adaptive Arc Subdivision.}
Between consecutive keypoints, we analyze the local geometry:
\begin{itemize}[nosep]
    \item If the angle between adjacent segments indicates a \emph{sharp corner} (acute angle), the keypoint is kept as-is.
    \item If the segment between keypoints is \emph{curved} (obtuse angle), we subdivide along the \emph{actual contour path} (not a straight line) by walking along the contour pixels and sampling at regular arc-length intervals.
\end{itemize}
The maximum arc-length between consecutive vertices is set to $\ell_{\max} = P / n_{\text{target}}$, where $n_{\text{target}} = 50$ is the target number of exterior vertices.
This ensures uniform vertex density along curves while preserving structural corners.
Figure~\ref{fig:contour_steps} illustrates this four-step extraction process.

\subsection{Interior Boundary Detection}
\label{sec:interior}

Unlike the exterior contour, which follows the mask boundary, interior vertices must be placed along visual features \emph{within} the sprite to enable independent deformation of visually distinct regions.
Our approach detects these internal boundaries using classical edge detection and places vertices along them.

\paragraph{Step 1: Bilateral Filtering.}
The input RGB image is filtered with a bilateral filter~\cite{tomasi1998bilateral} ($d = 9$, $\sigma_{\text{color}} = 75$, $\sigma_{\text{space}} = 75$) to suppress texture noise while preserving sharp edges between distinct visual regions.
This is critical for sprites, which often contain detailed textures that would produce excessive spurious edges.

\paragraph{Step 2: Multi-Channel Canny Edge Detection.}
We apply the Canny edge detector~\cite{canny1986computational} independently to each color channel (R, G, B) and to the grayscale image, using thresholds $t_{\text{low}} = 40$ and $t_{\text{high}} = 120$.
The per-channel edge maps are combined via bitwise OR:
\begin{equation}
    E = E_{\text{gray}} \cup E_R \cup E_G \cup E_B
\end{equation}
This multi-channel approach captures edges that are visible in only one or two color channels (\eg, a red outline on a blue background), which single-channel Canny would miss.

\paragraph{Step 3: Masking and Morphology.}
The edge map is masked by an eroded version of the binary mask (using an $11 \times 11$ kernel with 2 iterations) to exclude edges near the outer silhouette, which are already covered by the exterior contour vertices.
A morphological closing operation ($ 3 \times 3$ kernel) connects small gaps in broken edge segments.

\paragraph{Step 4: Contour-Aware Vertex Placement.}
Connected contours are extracted from the masked edge map, and those shorter than $\ell_{\min} = 150$ pixels are discarded.
For each remaining contour, we apply Douglas-Peucker simplification with $\epsilon_{\text{int}} = 0.012 \times P_c$ (where $P_c$ is the contour's arc length) to identify structural keypoints.
Between consecutive keypoints, we subdivide segments longer than $\ell_{\text{seg}} = 40$ pixels by sampling along the actual contour path.
Each candidate vertex is checked against the distance transform of the mask: only points at least $d_{\min} = 6$ pixels from the mask boundary are retained.

\begin{figure}[H]
    \centering
    \includegraphics[width=\columnwidth,height=0.18\textheight,keepaspectratio]{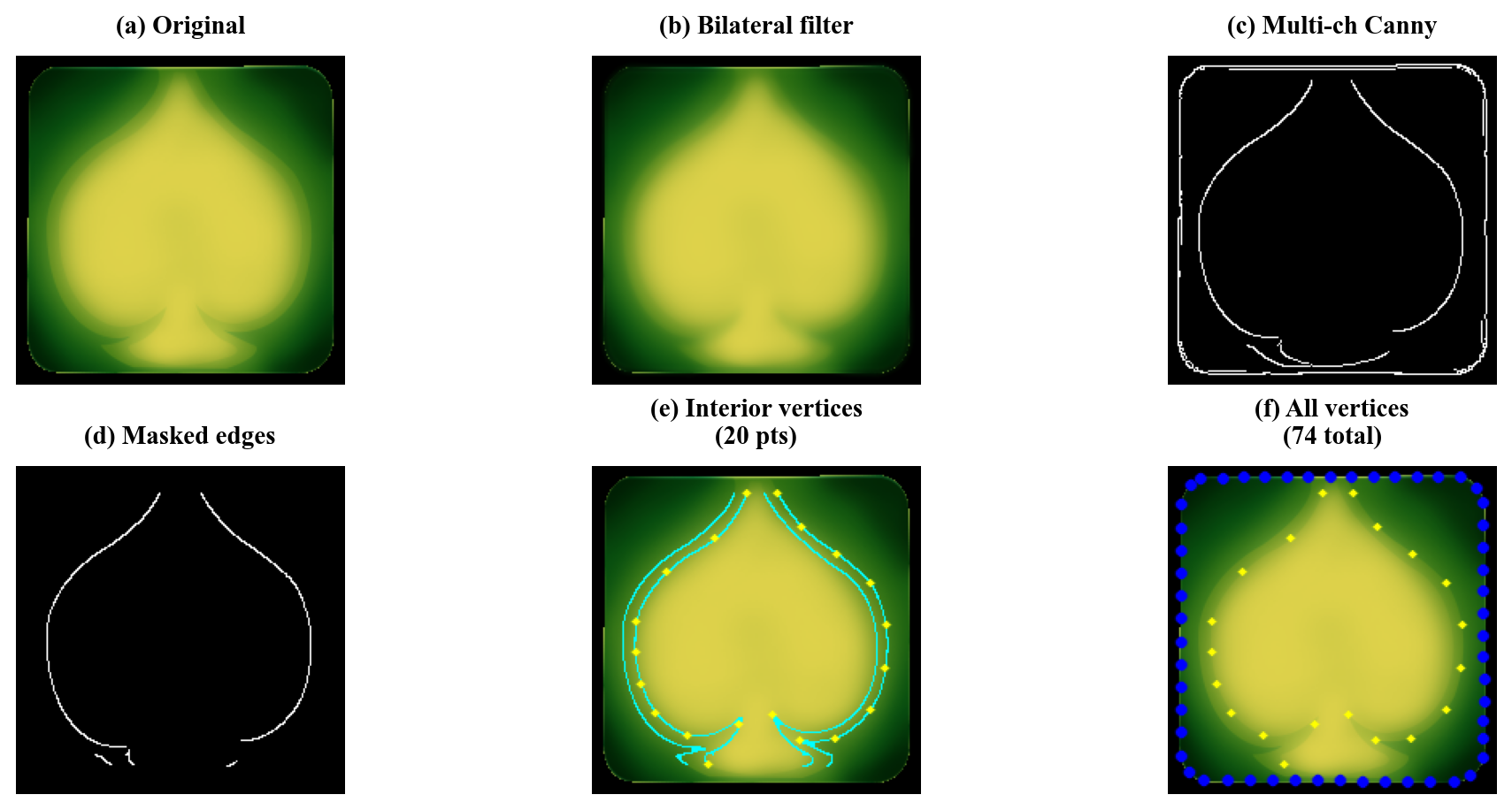}
    \caption{Interior vertex placement pipeline shown as a $2 \times 3$ grid. (a)~Original sprite. (b)~Bilateral-filtered image (texture noise reduced, structural edges preserved). (c)~Multi-channel Canny edge map (union of per-channel detections). (d)~Masked edges after erosion and morphological closing. (e)~Interior vertices (yellow) placed along surviving contours using Douglas-Peucker simplification and uniform subdivision. (f)~All vertices combined: exterior (blue) and interior (yellow), ready for Delaunay triangulation.}
    \label{fig:vertex_detail}
\end{figure}

\paragraph{Step 5: Deduplication.}
All interior vertices are globally deduplicated with a minimum distance threshold of $r = 18$ pixels to prevent excessive vertex density in regions where multiple detected contours converge.
Figure~\ref{fig:vertex_detail} visualizes the complete interior vertex placement pipeline.

\subsection{Triangulation and Filtering}
\label{sec:triangulation}

The exterior and interior vertices are combined and triangulated using Delaunay triangulation~\cite{delaunay1934sphere} via OpenCV's \texttt{Subdiv2D}.
Duplicate vertices (within 2 pixels) are removed before triangulation.
Each resulting triangle is evaluated by computing its centroid; triangles whose centroid falls outside the binary mask $M$ are discarded.
This ensures the final mesh covers only the sprite foreground.

The output is a triangle mesh $(V, T)$ where $V \in \mathbb{R}^{N \times 2}$ are vertex positions and $T \in \mathbb{N}^{K \times 3}$ are triangle indices, exported in JSON format compatible with Spine2D.

\section{Dataset}
\label{sec:dataset}

\subsection{Data Collection}

To train our segmentation network, we collected a large-scale dataset of 2D sprites from 172 distinct games that use the Spine2D animation framework.
For each game, we gathered the complete set of Spine assets: skeleton definition files (\texttt{.skel} binary or \texttt{.json}), texture atlases (\texttt{.atlas}), and sprite sheet images (\texttt{.png}, \texttt{.avif}, \texttt{.webp}).
The data preparation pipeline consists of three main stages: asset collection, binary skeleton conversion, and sample extraction.

\paragraph{Stage 1: Asset Collection.}
We collected Spine2D asset packages from 172 distinct web-based games that use the framework.
For each game, we gathered the complete set of runtime assets: skeleton definition files (\texttt{.skel} or \texttt{.json}), texture atlas descriptors (\texttt{.atlas}), and sprite sheet images (\texttt{.png}, \texttt{.avif}, \texttt{.webp}).
Asset files were organized by game and deduplicated based on content hashes.

\paragraph{Stage 2: Binary Skeleton Conversion with SkelToJson.}
Spine2D skeleton files are distributed in two formats: a human-readable JSON format and a compact binary \texttt{.skel} format optimized for runtime loading.
The binary format is a densely packed byte stream that encodes the entire skeleton hierarchy (bones, slots, skins, attachments, animations, constraints) in a format not directly readable.
Critically, the mesh vertex coordinates, UV mappings, and triangle indices needed for our dataset are embedded within these binary files.

\begin{figure}[H]
    \centering
    \includegraphics[width=\columnwidth,height=0.16\textheight,keepaspectratio]{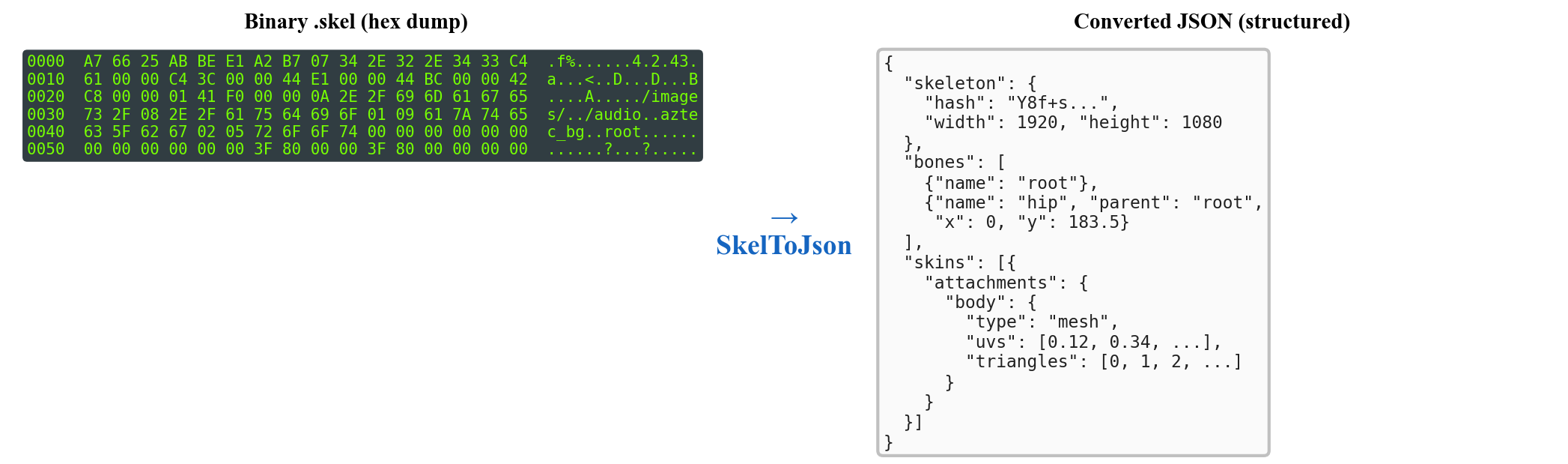}
    \caption{Example of binary-to-JSON conversion using SkelToJson. The binary \texttt{.skel} format encodes the complete skeleton as a dense byte stream (left). Our converter produces structured JSON (right) from which mesh vertex positions, UV coordinates, and triangle indices can be directly extracted.}
    \label{fig:skel_to_json}
\end{figure}

To decode the binary format at scale, we developed \textsc{SkelToJson}~\cite{skelToJson}, an open-source Python library that performs lossless conversion of Spine \texttt{.skel} binary files to structured JSON.
The converter reverse-engineers the Spine 4.2 and 4.3 binary protocols, implementing a sequential byte reader that parses:
\begin{itemize}[nosep]
    \item \textbf{Skeleton metadata}: version hash, dimensions, reference scale;
    \item \textbf{Bone hierarchy}: parent-child relationships, positions, rotations, scales;
    \item \textbf{Slots and skins}: draw order, attachment mappings across skin variants;
    \item \textbf{Attachments}: region attachments (simple rectangles), mesh attachments (with vertices, UVs, triangle indices, edge data, and deformation weights), bounding boxes, paths, and clipping regions;
    \item \textbf{Constraints}: IK, transform, path, and physics constraints;
    \item \textbf{Animations}: keyframed timelines for all animated properties.
\end{itemize}
We applied SkelToJson to all 882 binary \texttt{.skel} files in our collection, achieving a 100\% conversion success rate.
The library is released publicly as a PyPI package (\texttt{pip install SkelToJson}) and on GitHub\footnote{\url{https://github.com/BastienGimbert/SkelToJson}}.
Figure~\ref{fig:skel_to_json} illustrates an example of this conversion.

\begin{figure}[H]
    \centering
    \includegraphics[width=\columnwidth]{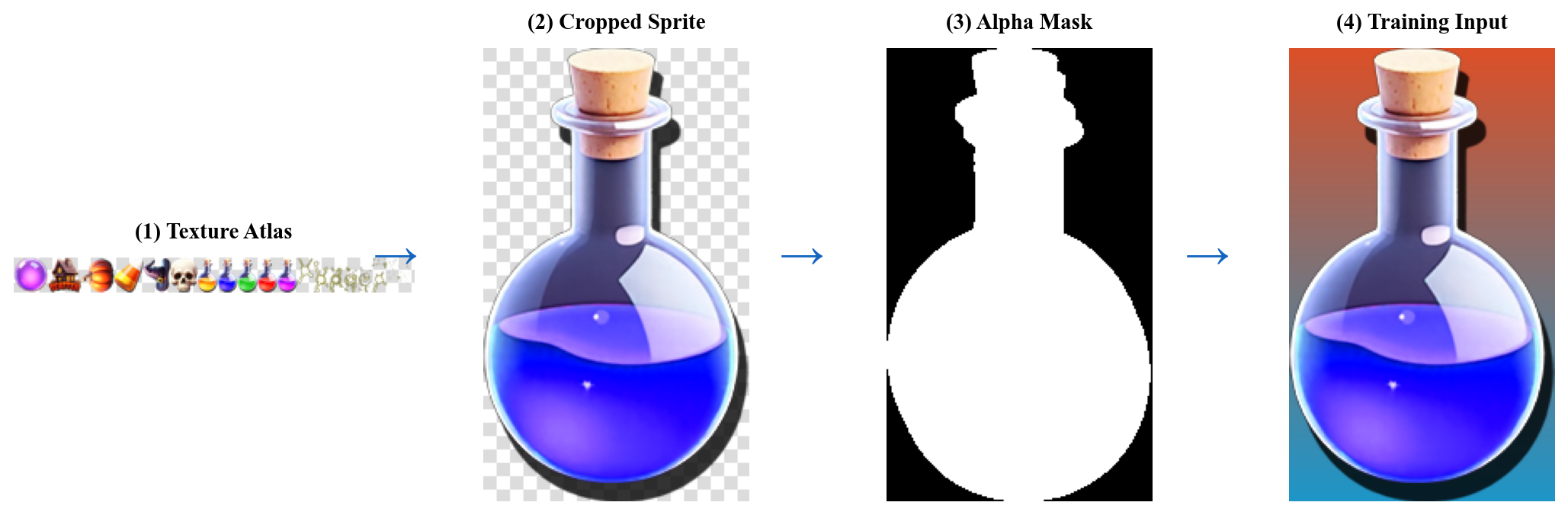}
    \caption{Dataset preparation pipeline shown in four panels. (1)~A full texture atlas from a game. (2)~An individual sprite attachment cropped from the atlas using region definitions parsed from the \texttt{.atlas} file. (3)~The binary mask extracted from the sprite's alpha channel ($\alpha > 128$). (4)~The sprite composited onto a random gradient background, as used during training to force the network to learn foreground--background separation from visual features rather than trivial alpha thresholding.}
    \label{fig:dataset_pipeline}
\end{figure}

\paragraph{Stage 3: Dataset Preparation Pipeline.}
From the converted JSON skeleton files and their associated texture atlases, we extract individual sprite attachments as training samples.
Our preparation pipeline performs the following steps for each project:
\begin{enumerate}[nosep]
    \item \textbf{Atlas parsing}: We parse \texttt{.atlas} files to extract region definitions (bounding boxes, rotation flags, scale factors) for each named sprite within the texture sheet.
    \item \textbf{Skeleton loading}: Skeleton data is loaded from JSON or converted from binary \texttt{.skel} using SkelToJson.
    \item \textbf{Attachment extraction}: For each skin in the skeleton, we iterate over all slots and extract attachment definitions. For \emph{region} attachments, we extract position, rotation, and scale relative to the parent bone. For \emph{mesh} attachments, we additionally extract vertex positions, UV coordinates, triangle indices, and edge vertex lists.
    \item \textbf{Image cropping}: Each attachment's image region is cropped from its parent texture atlas, handling rotation, whitespace stripping, and multi-resolution atlases. The result is a clean RGBA image of the individual sprite.
    \item \textbf{Ground truth generation}: The alpha channel provides the segmentation mask. For mesh attachments, UV coordinates are mapped to pixel space to produce vertex position annotations.
\end{enumerate}
Figure~\ref{fig:dataset_pipeline} illustrates this pipeline from atlas to training sample.

\subsection{Dataset Statistics}

The resulting dataset contains \textbf{100,363} samples from 172 games:
\begin{itemize}[nosep]
    \item \textbf{74,366} ``region'' samples with binary masks only;
    \item \textbf{25,997} ``mesh'' samples with both masks and vertex annotations.
\end{itemize}

\begin{figure}[H]
    \centering
    \includegraphics[width=\columnwidth,height=0.25\textheight,keepaspectratio]{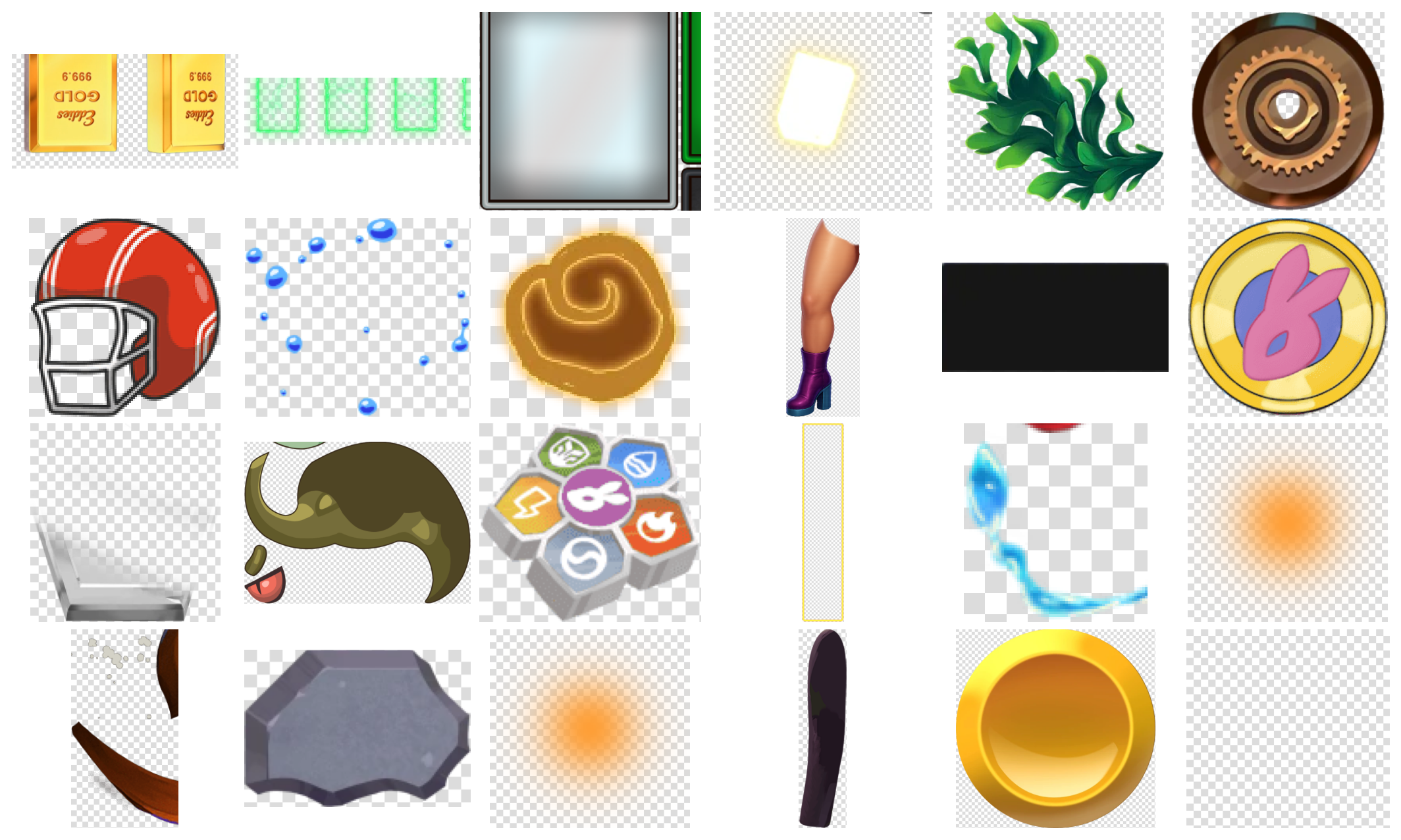}
    \caption{Representative samples from our dataset, shown as a $4 \times 6$ grid where each sprite is drawn from a different game. The dataset spans 172 games and exhibits wide diversity in artistic style, subject matter, complexity, and resolution---including characters, objects, UI elements, effects, and accessories.}
    \label{fig:dataset_samples}
\end{figure}

\begin{figure}[H]
    \centering
    \includegraphics[width=\columnwidth,height=0.18\textheight,keepaspectratio]{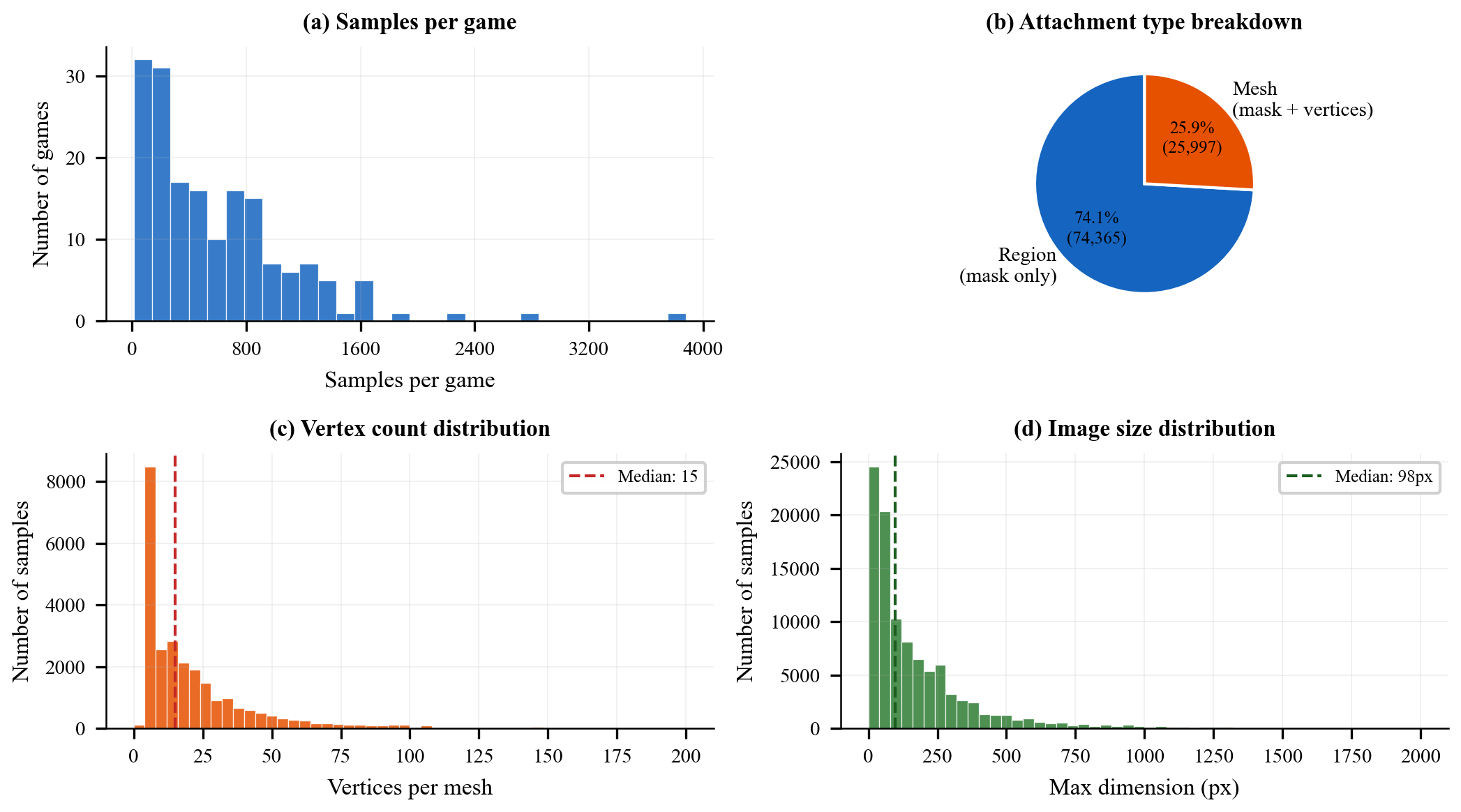}
    \caption{Dataset statistics shown as four panels. (a)~Histogram of the number of samples per game. (b)~Pie chart of attachment type breakdown: region (mask only) vs.\ mesh (mask + vertex annotations) with percentages and counts. (c)~Distribution of vertex counts for mesh samples, with a median line. (d)~Distribution of max image dimension (in pixels) across all samples, with a median line.}
    \label{fig:dataset_stats}
\end{figure}

The sprites exhibit wide diversity in artistic style, complexity, resolution, and subject matter (characters, UI elements, effects, backgrounds).
Figure~\ref{fig:dataset_samples} shows representative samples from the dataset illustrating this diversity.
We split the dataset into training (90\%) and validation (10\%) sets, stratified by project to prevent data leakage.
Figure~\ref{fig:dataset_stats} provides detailed breakdowns of the dataset composition.

\subsection{Data Augmentation}

\begin{figure}[H]
    \centering
    \includegraphics[width=\columnwidth]{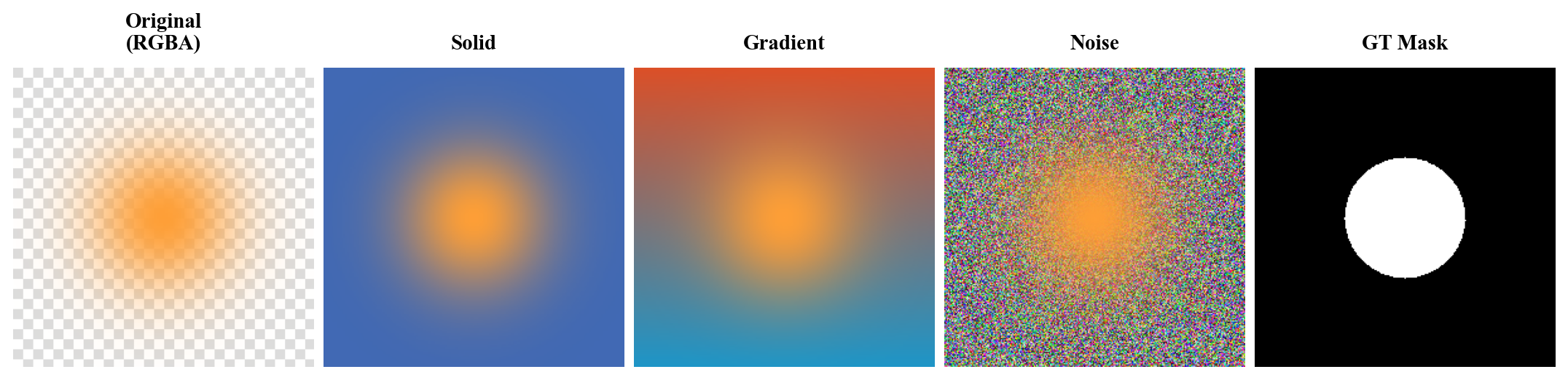}
    \caption{Data augmentation via background composition, shown as five horizontal panels. From left to right: the original RGBA sprite on a checkerboard (showing transparency), then the same sprite composited onto a solid color, a gradient, and a noise pattern. The rightmost panel shows the ground truth binary mask derived from the alpha channel. This augmentation forces the segmentation network to learn foreground--background separation from visual features.}
    \label{fig:augmentation}
\end{figure}

During training, each sprite is composited onto a randomly generated background to simulate the absence of alpha channels:
\begin{itemize}[nosep]
    \item Solid color (random RGB);
    \item Gradient (random direction and colors);
    \item Noise (Perlin-like patterns).
\end{itemize}
This forces the segmentation network to learn foreground--background separation from visual features rather than relying on a trivial alpha threshold, making it robust to diverse deployment scenarios.
Figure~\ref{fig:augmentation} illustrates these augmentation variants.

All images are resized to $256 \times 256$ and normalized with ImageNet statistics ($\mu = [0.485, 0.456, 0.406]$, $\sigma = [0.229, 0.224, 0.225]$).

\section{Experiments}
\label{sec:experiments}

\subsection{Segmentation Performance}

We train our EfficientNet--U-Net segmentation model for 30 epochs on the full dataset (100,363 samples) using a batch size of 32 on a single NVIDIA RTX 4070.

Table~\ref{tab:seg_results} reports the segmentation performance on the validation set at the best epoch (epoch 14).

\begin{table}[H]
    \centering
    \caption{Segmentation performance on the validation set.}
    \label{tab:seg_results}
    \begin{tabular}{lc}
        \toprule
        \textbf{Metric} & \textbf{Value} \\
        \midrule
        IoU & 0.8697 \\
        Dice Score & 0.9304 \\
        Pixel Accuracy & 0.9741 \\
        BCE Loss & 0.0523 \\
        Dice Loss & 0.0696 \\
        \bottomrule
    \end{tabular}
\end{table}

\begin{figure}[H]
    \centering
    \includegraphics[height=0.18\textheight,keepaspectratio]{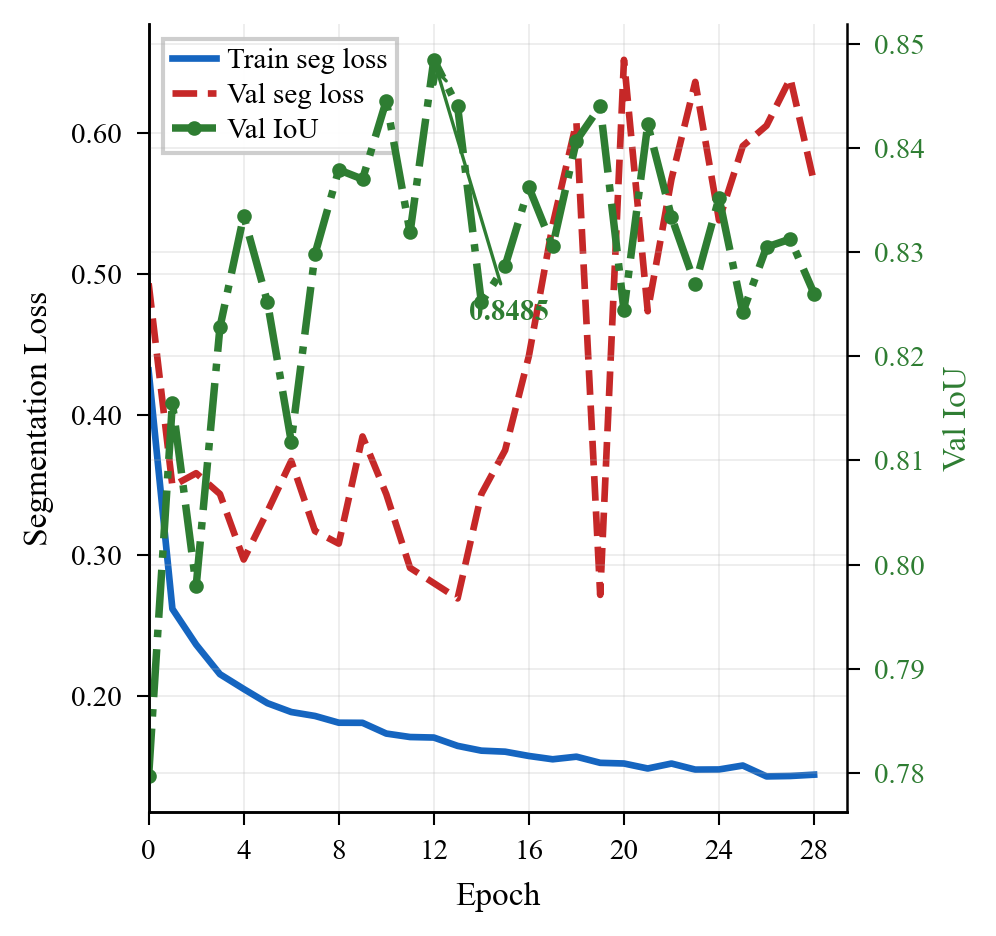}
    \caption{Training dynamics shown on a single dual-axis plot. The left axis displays the segmentation loss (BCE + Dice) for both training and validation sets, which decrease steadily with a minimal overfitting gap. The right axis shows the validation IoU, which increases during training and reaches its best value, after which overfitting begins.}
    \label{fig:training_curves}
\end{figure}

\begin{figure}[H]
    \centering
    \includegraphics[width=\columnwidth,height=0.18\textheight,keepaspectratio]{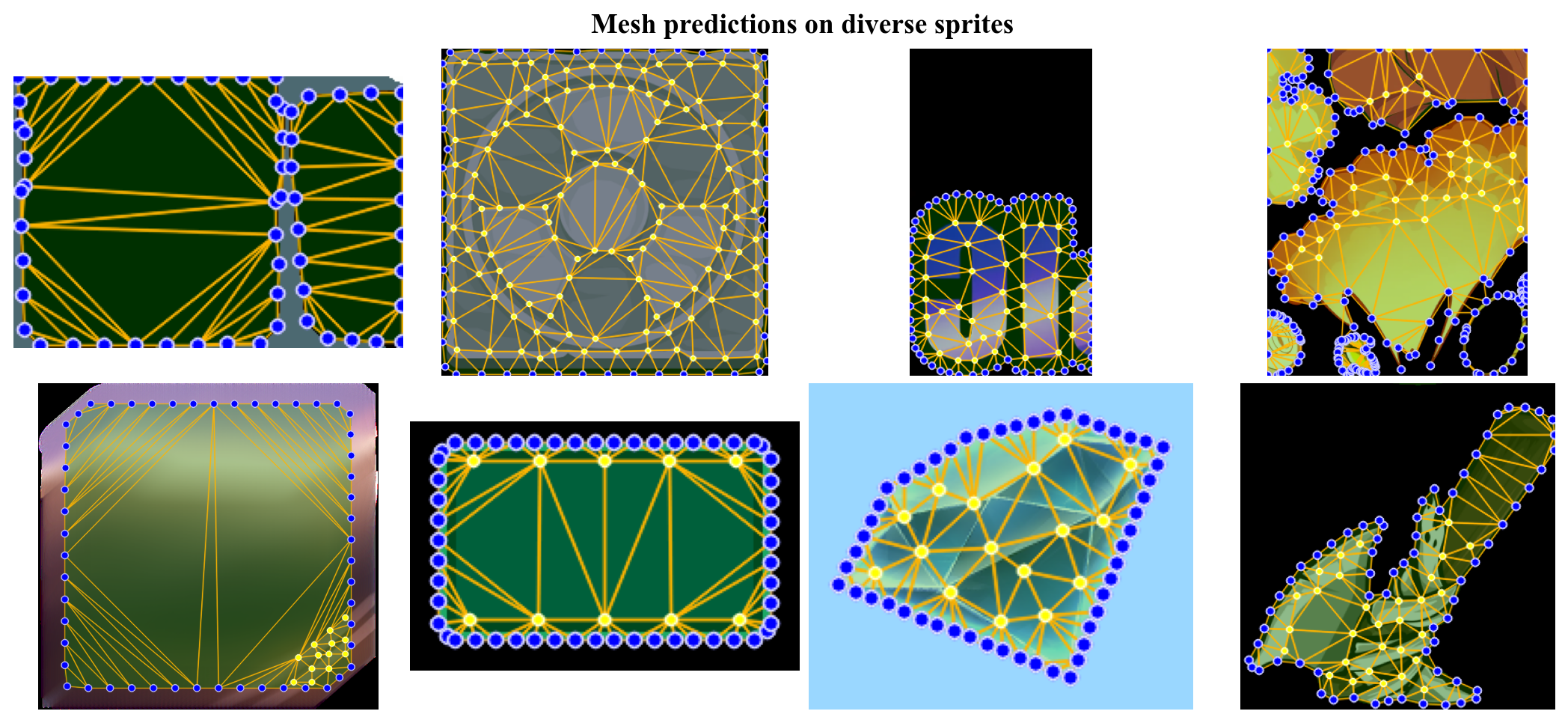}
    \caption{Mesh prediction results on eight diverse sprites arranged in a $2 \times 4$ grid. Each cell shows a sprite with the automatically generated mesh overlaid as a wireframe, demonstrating that our pipeline handles a variety of art styles, shapes, and complexities.}
    \label{fig:seg_success_failure}
\end{figure}

The model achieves strong segmentation performance across diverse sprite styles.
Failure cases are concentrated on sprites with extremely thin structures (\eg, hair strands, weapon tips) and sprites whose foreground is very similar in color to the composited background.
Figure~\ref{fig:training_curves} shows the training dynamics, and Figure~\ref{fig:seg_success_failure} presents qualitative mesh results on diverse sprites.

\begin{figure}[H]
    \centering
    \includegraphics[width=\columnwidth,height=0.22\textheight,keepaspectratio]{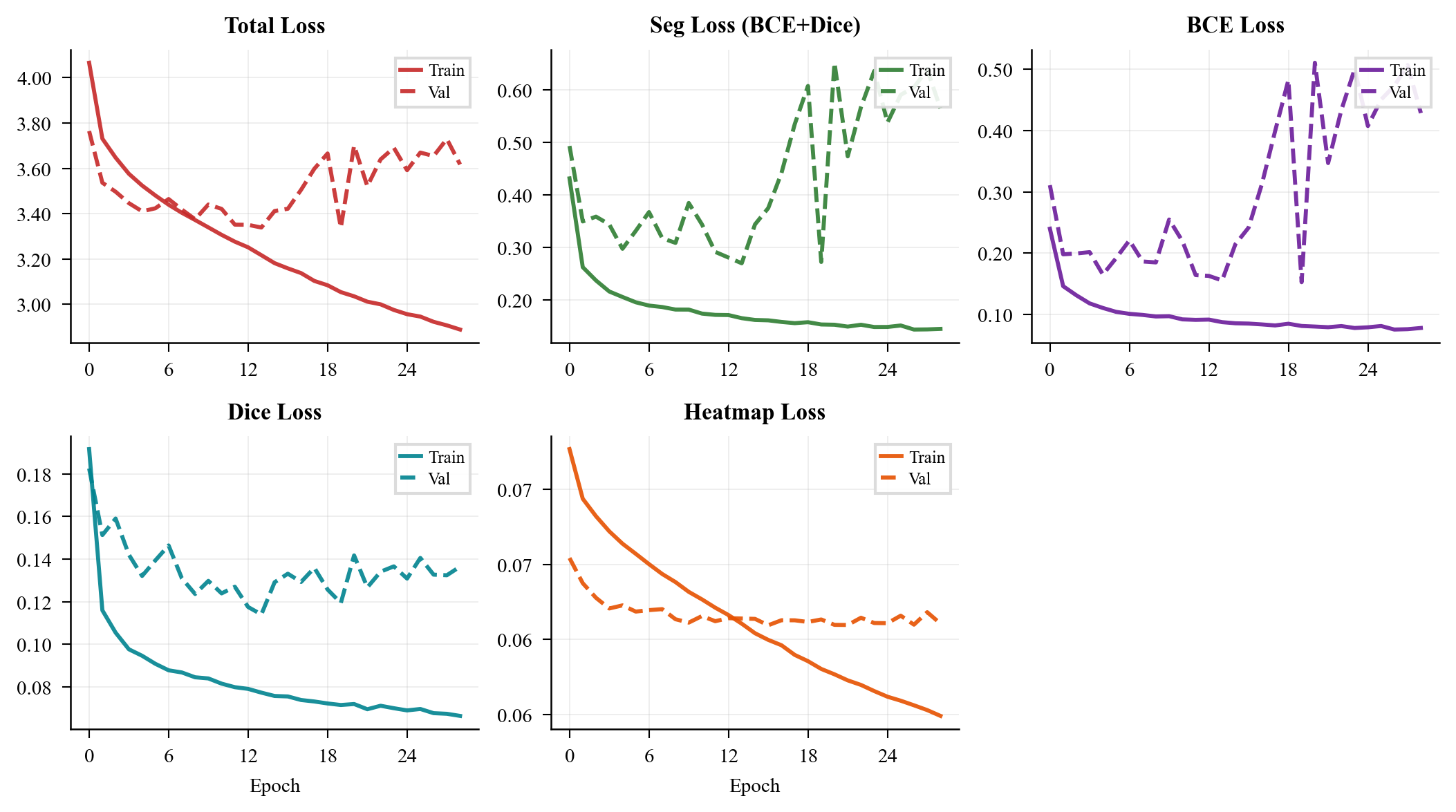}
    \caption{Detailed training loss decomposition shown as a $2 \times 3$ grid (five panels, one empty). Each panel plots training and validation curves for a single loss component: total loss, segmentation loss (BCE + Dice combined), BCE loss, Dice loss, and heatmap MSE loss. The segmentation-related losses decrease steadily on both sets, while the heatmap loss plateaus early, confirming the infeasibility of direct vertex prediction.}
    \label{fig:seg_predictions}
\end{figure}

\subsection{Heatmap Regression Failure}

As discussed in Section~\ref{sec:heatmap_failure}, the vertex heatmap regression head failed to converge.
Table~\ref{tab:heatmap_failure} summarizes the comparison between the segmentation and heatmap decoder heads during training.

\begin{table}[H]
    \centering
    \caption{Comparison of segmentation vs.\ heatmap decoder convergence. The heatmap head plateaus without producing meaningful predictions.}
    \label{tab:heatmap_failure}

    \resizebox{\columnwidth}{!}{%
    \begin{tabular}{lcc}
        \toprule
        & \textbf{Segmentation} & \textbf{Heatmap} \\
        \midrule
        Final loss & 0.122 & 0.061 (plateau) \\
        IoU / Peak recall & 0.87 & $\approx$ 0 \\
        Converged? & \checkmark & $\times$ \\
        Prediction quality & Accurate masks & Uniform blur \\
        \bottomrule
    \end{tabular}%
    }
\end{table}

\begin{figure}[H]
    \centering
    \includegraphics[width=\columnwidth,height=0.16\textheight,keepaspectratio]{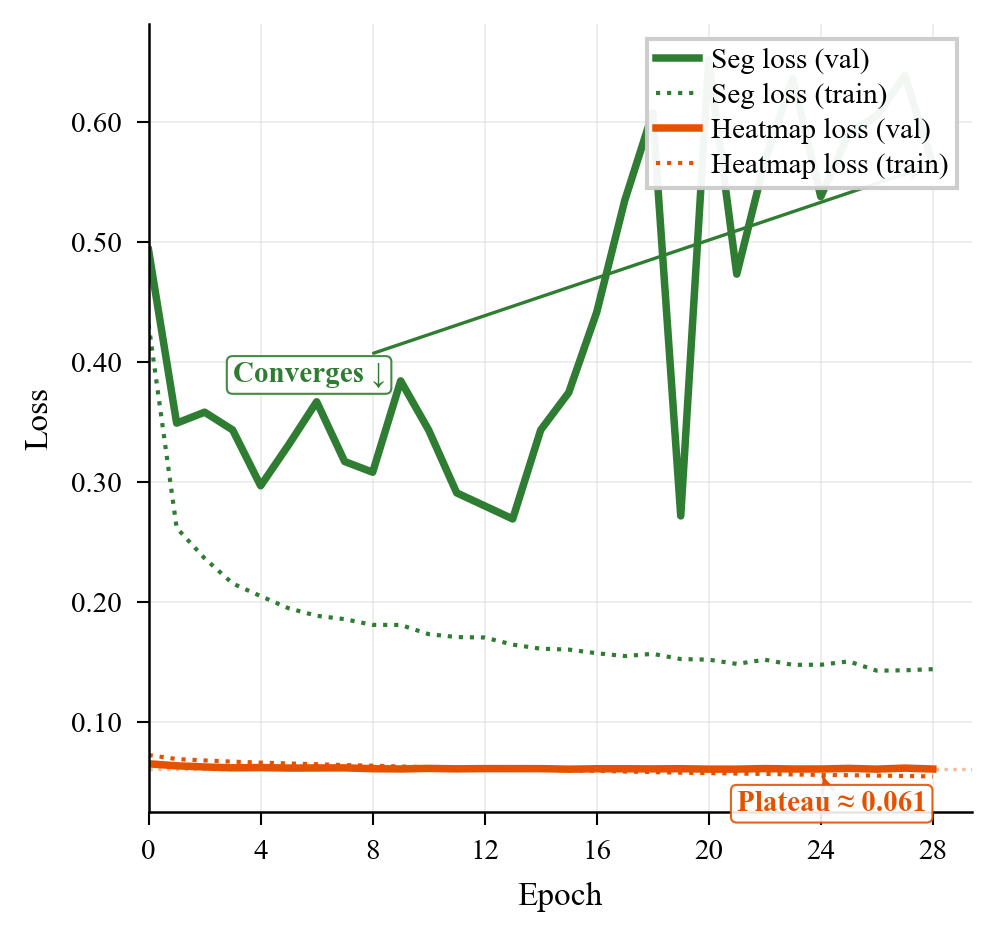}
    \caption{Training loss comparison between segmentation and heatmap decoder heads. Four curves are shown: segmentation loss (green, train and val) and heatmap loss (orange, train and val). The segmentation loss decreases steadily throughout training, while the heatmap loss plateaus at approximately 0.061 after the first few epochs and never improves, demonstrating the fundamental non-learnability of vertex positions from pixel values alone.}
    \label{fig:heatmap_loss}
\end{figure}

\begin{figure}[H]
    \centering
    \includegraphics[width=\columnwidth,height=0.16\textheight,keepaspectratio]{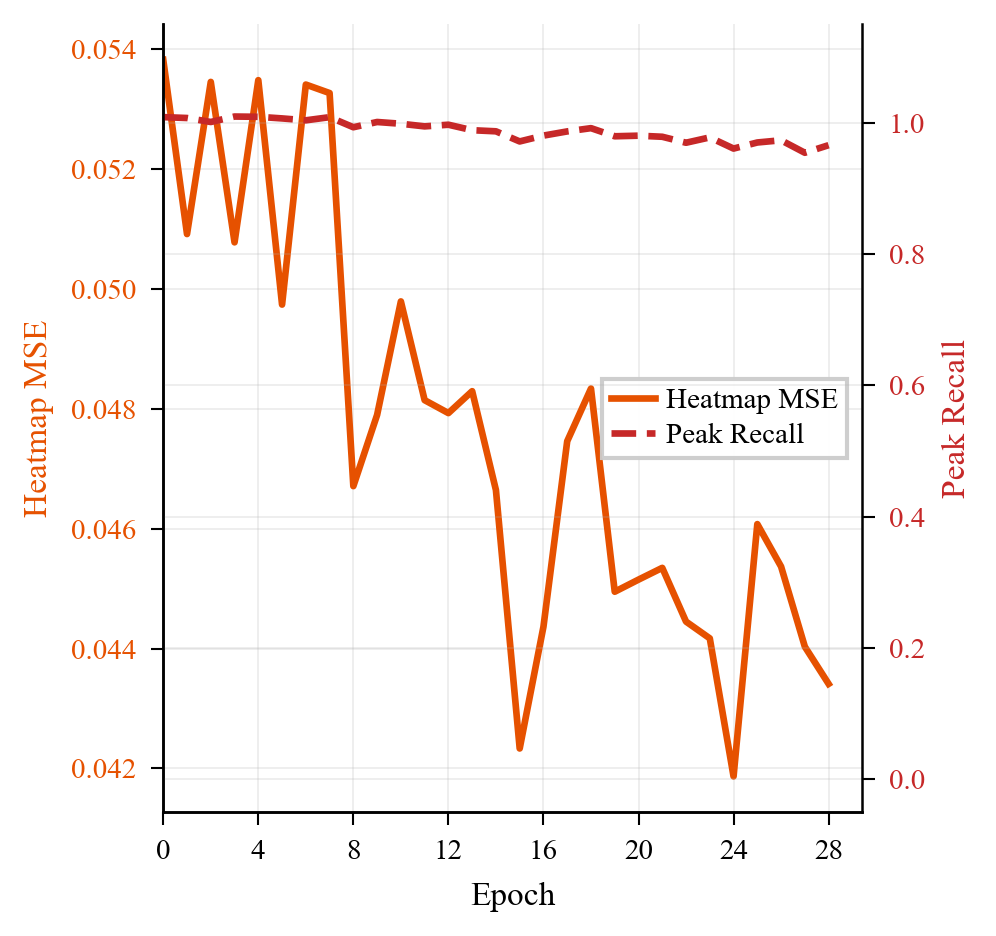}
    \caption{Heatmap decoder failure analysis. The heatmap MSE (left axis) remains elevated throughout training with no convergence trend, while the peak recall metric (right axis) drops from an initial value near 1.0, indicating the network's inability to localize vertex positions. This contrasts sharply with the segmentation decoder's steady improvement under identical training conditions.}
    \label{fig:heatmap_visual}
\end{figure}

This result confirms that vertex placement for animation meshes cannot be formulated as a straightforward regression problem from pixel values.
Figure~\ref{fig:heatmap_loss} visualizes the training loss comparison between the two decoder heads, and Figure~\ref{fig:heatmap_visual} further illustrates the heatmap decoder's inability to localize vertices.
A detailed per-component loss decomposition is provided in Figure~\ref{fig:seg_predictions}.

\subsection{Mesh Generation Quality}

We evaluate the quality of generated meshes on a held-out test set of diverse sprite images.
Table~\ref{tab:mesh_stats} reports mesh statistics.

\begin{table}[H]
    \centering
    \caption{Mesh generation statistics across test sprites. Values reported as mean $\pm$ std.}
    \label{tab:mesh_stats}
    \begin{tabular}{lc}
        \toprule
        \textbf{Metric} & \textbf{Value} \\
        \midrule
        Exterior vertices & $64 \pm 12$ \\
        Interior vertices & $420 \pm 180$ \\
        Total vertices & $484 \pm 185$ \\
        Triangles & $890 \pm 350$ \\
        Processing time & $1.8 \pm 0.6$ s \\
        Mask coverage & $> 99.5$\% \\
        \bottomrule
    \end{tabular}
\end{table}

\begin{figure}[H]
    \centering
    \includegraphics[width=\columnwidth,height=0.3\textheight,keepaspectratio]{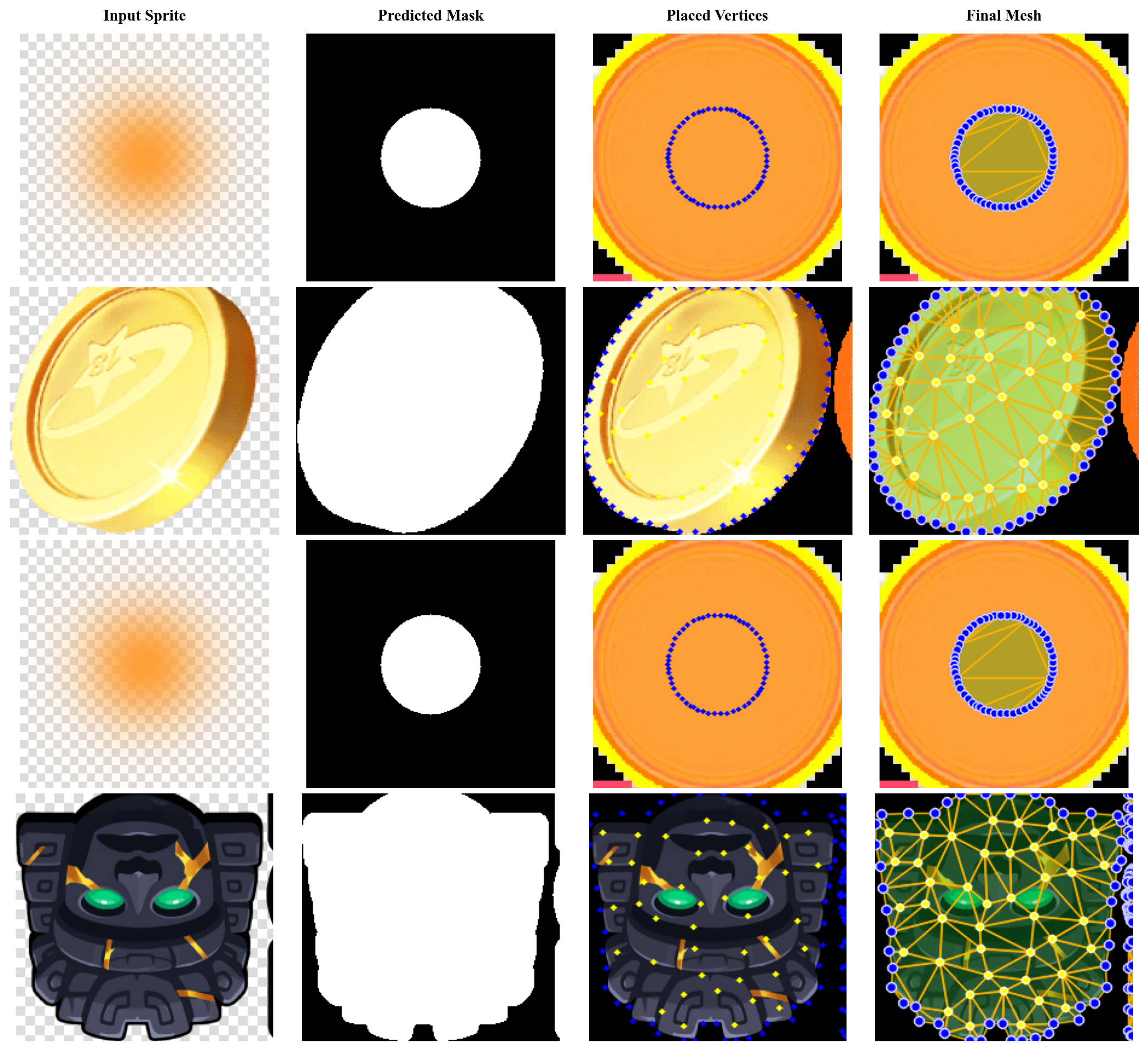}
    \caption{Qualitative results on four diverse test sprites, arranged with one sprite per row. Each row shows four panels from left to right: input sprite image, predicted binary segmentation mask, detected vertices (blue = exterior contour, yellow = interior boundary), and final triangulated mesh overlaid on the sprite. Interior vertices are placed along detected visual boundaries within the sprite.}
    \label{fig:qualitative}
\end{figure}

\begin{figure}[H]
    \centering
    \includegraphics[width=\columnwidth,height=0.16\textheight,keepaspectratio]{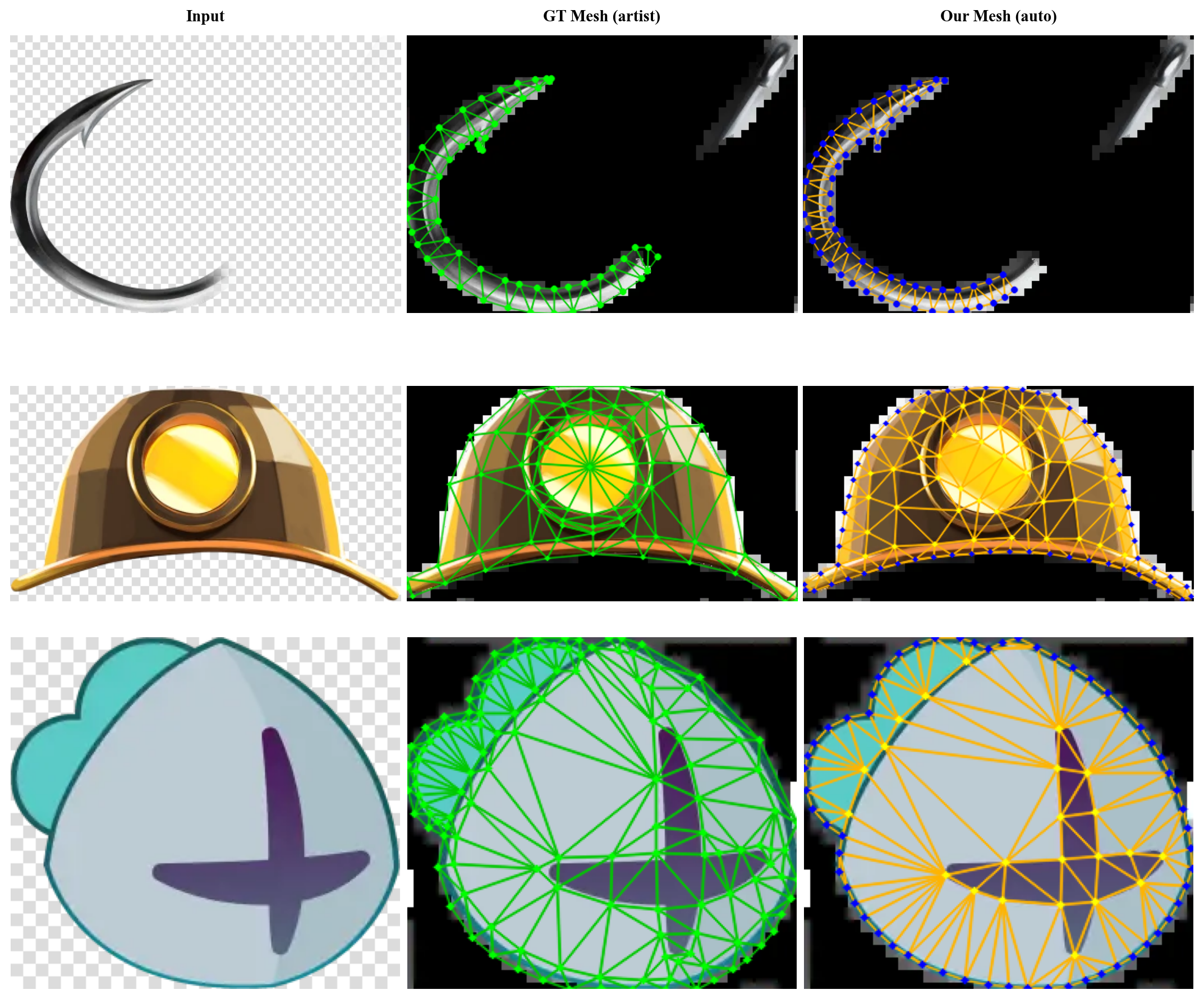}
    \caption{Comparison of artist-created and automatically generated meshes, shown as three columns per row: input sprite (left), ground truth artist mesh in green (center), and our predicted mesh in orange with blue exterior and yellow interior vertices (right). While the exact vertex positions differ---reflecting the subjective nature of the task---both meshes cover similar visual boundaries. Our method tends to produce denser meshes along detected edges, while artists place vertices more selectively based on anticipated animation needs.}
    \label{fig:gt_vs_pred}
\end{figure}

Figure~\ref{fig:qualitative} provides a detailed step-by-step visualization of our pipeline on diverse test sprites.
Figure~\ref{fig:gt_vs_pred} compares our automatically generated meshes against artist-created ground truth, illustrating that while vertex positions differ, both approaches capture similar visual boundaries.

\subsection{Ablation Study}
\label{sec:ablation}

We conduct ablation experiments to evaluate the contribution of each component of our interior vertex placement pipeline.
We measure mesh quality using three criteria: (1)~visual boundary adherence (percentage of significant internal edges that have at least one vertex within 10 pixels), (2)~vertex efficiency (number of vertices needed to achieve a given coverage), and (3)~triangle regularity (ratio of minimum to maximum angles per triangle, averaged over all triangles).

\begin{table}[H]
    \centering
    \caption{Ablation study. We disable individual components and measure their impact on mesh quality.}
    \label{tab:ablation}

    \resizebox{\columnwidth}{!}{%
    \begin{tabular}{lccc}
        \toprule
        \textbf{Configuration} & \textbf{Boundary} & \textbf{Vertices} & \textbf{Angle} \\
        & \textbf{Adherence} & \textbf{Count} & \textbf{Ratio} \\
        \midrule
        Full pipeline & \textbf{78.3\%} & 484 & \textbf{0.42} \\
        \midrule
        No bilateral filter & 71.2\% & 623 & 0.38 \\
        Single-channel Canny & 64.8\% & 412 & 0.41 \\
        No DP simplification & 76.5\% & 1,847 & 0.31 \\
        No arc subdivision & 73.1\% & 389 & 0.40 \\
        Grid interior only & 42.6\% & 520 & \textbf{0.42} \\
        No interior vertices & 0\% & 64 & 0.39 \\
        \bottomrule
    \end{tabular}%
    }
\end{table}


Key findings:
\begin{itemize}[nosep]
    \item \textbf{Bilateral filtering} reduces spurious vertices by 22\% while improving boundary adherence by 7.1 percentage points, confirming its importance for noise reduction.
    \item \textbf{Multi-channel Canny} captures 13.5 more percentage points of boundaries than grayscale-only detection, demonstrating that color-space edges carry significant information.
    \item \textbf{Douglas-Peucker simplification} is essential for vertex efficiency: without it, 3.8$\times$ more vertices are generated with only marginally better boundary adherence, severely degrading triangle quality.
    \item \textbf{Grid-based interior placement} achieves good triangle regularity but fails to follow visual boundaries, confirming the need for contour-aware placement.
\end{itemize}

\subsection{Comparison with Baselines}

\begin{figure}[H]
    \centering
    \includegraphics[width=\columnwidth,height=0.22\textheight,keepaspectratio]{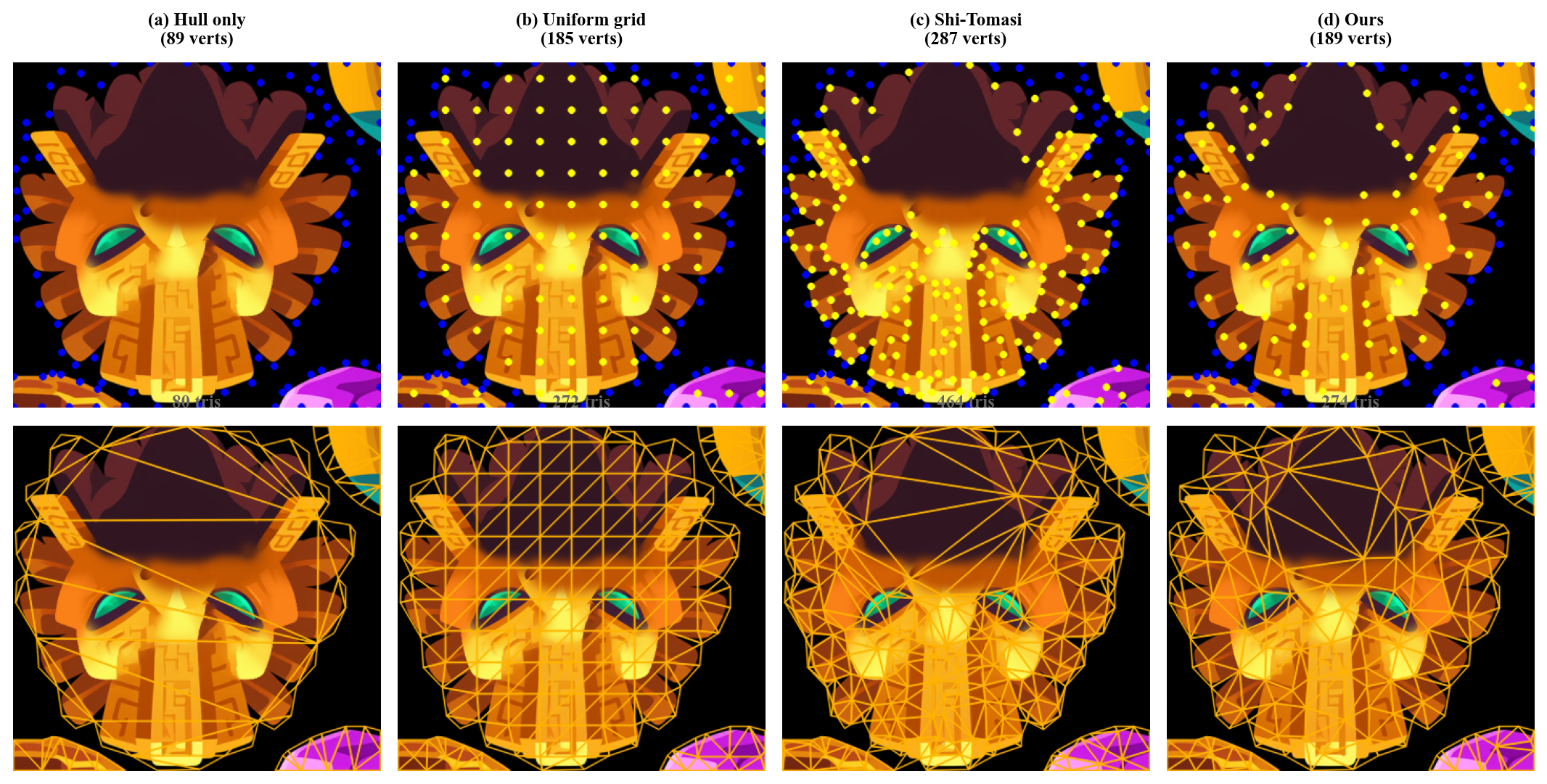}
    \caption{Comparison of four mesh generation approaches on the same sprite, shown as a $2 \times 4$ grid. Top row: vertex placement (blue = exterior, yellow = interior) with vertex counts. Bottom row: resulting triangulated mesh wireframe with triangle counts. (a)~Hull only---exterior contour vertices with no interior, yielding a single deformable region. (b)~Uniform grid---regular interior vertices that ignore visual structure. (c)~Shi-Tomasi corners---feature points that do not follow edge contours. (d)~Our contour-aware method---vertices placed along internal visual boundaries, enabling independent deformation of visually distinct regions.}
    \label{fig:comparison}
\end{figure}

We compare \method{} against three baseline approaches for interior vertex placement:

\begin{enumerate}[nosep]
    \item \textbf{Alpha-only hull}: Vertices placed only along the outer contour with no interior vertices. This is analogous to the simplest auto-mesh tools.
    \item \textbf{Uniform grid}: Interior vertices placed on a regular grid with spacing proportional to the mask size, filtered by mask membership.
    \item \textbf{Shi-Tomasi corners}~\cite{shi1994good}: Interior vertices placed at Shi-Tomasi corner features detected in the sprite image.
\end{enumerate}
Figure~\ref{fig:comparison} provides a visual comparison of all four approaches on the same sprite.

\begin{table}[H]
    \centering
    \caption{Comparison with baseline methods.}
    \label{tab:comparison}

    \resizebox{\columnwidth}{!}{%
    \begin{tabular}{lccc}
        \toprule
        \textbf{Method} & \textbf{Boundary} & \textbf{Vertices} & \textbf{Articulable} \\
        & \textbf{Adherence} & & \textbf{Regions} \\
        \midrule
        Alpha-only hull & 0\% & 64 & 1 \\
        Uniform grid & 42.6\% & 520 & Partial \\
        Shi-Tomasi corners & 55.3\% & 667 & Partial \\
        \method{} (ours) & \textbf{78.3\%} & 484 & \textbf{Full} \\
        \bottomrule
    \end{tabular}%
    }
\end{table}

Our method achieves the highest boundary adherence with a moderate vertex count, producing meshes where visually distinct regions can be independently articulated during animation.

\subsection{Runtime Performance}

The complete pipeline processes a single $6000 \times 6000$ pixel sprite in under 3 seconds on a consumer-grade GPU (NVIDIA RTX 4070 Laptop, 8 GB VRAM):
\begin{itemize}[nosep]
    \item Mask acquisition (alpha path): $< 0.1$ s
    \item Mask acquisition (model path): $\approx 0.3$ s
    \item Exterior contour extraction: $\approx 0.2$ s
    \item Interior boundary detection: $\approx 1.2$ s
    \item Delaunay triangulation: $\approx 0.1$ s
\end{itemize}

This represents a speedup of $300\times$--$1200\times$ compared to manual mesh creation (15--60 minutes).

\section{Limitations and Future Work}
\label{sec:limitations}

\paragraph{Parameter Sensitivity.}
Our interior detection pipeline involves several threshold parameters (Canny thresholds, minimum contour length, deduplication radius, \etc).
While the default values work well across diverse sprites, optimal settings may vary with artistic style and desired mesh density.
Automatic parameter tuning, perhaps conditioned on image statistics or sprite category, is a promising direction.

\paragraph{Semantic Awareness.}
Our method detects visual boundaries purely from low-level image features (edges, color transitions) without understanding the semantic content of the sprite (\eg, distinguishing arms from legs, or identifying clothing layers).
Integrating semantic segmentation or part-based decomposition could enable smarter vertex placement that aligns with functional anatomy.

\paragraph{Skeleton Generation.}
The natural extension of automatic mesh generation is automatic bone/skeleton placement.
A complete auto-rigging pipeline that produces both the mesh and the bone hierarchy from a single sprite image would fully automate the animation setup process.

\paragraph{Style Adaptation.}
Our segmentation model was trained exclusively on Spine2D sprites extracted from online casino and slot games.
Although the dataset spans 172 games with diverse artistic traditions (anime, cartoon, pixel art), all training data shares common characteristics: RGBA sprites with alpha-channel transparency, organic shapes designed for skeletal deformation, and textures typical of the gaming industry.
As a result, the model does \emph{not} generalize to arbitrary images such as logos, photographs, vector icons, or non-game artwork.
When applied to out-of-distribution inputs (\eg, a corporate logo with sharp geometric shapes and solid colors), the segmentation network produces unreliable masks, and the Canny-based interior vertex placement generates excessive spurious vertices along every color boundary.
Domain adaptation or fine-tuning on a broader dataset would be necessary to extend the method beyond Spine2D game assets.

\begin{figure}[H]
    \centering
    \includegraphics[width=\columnwidth]{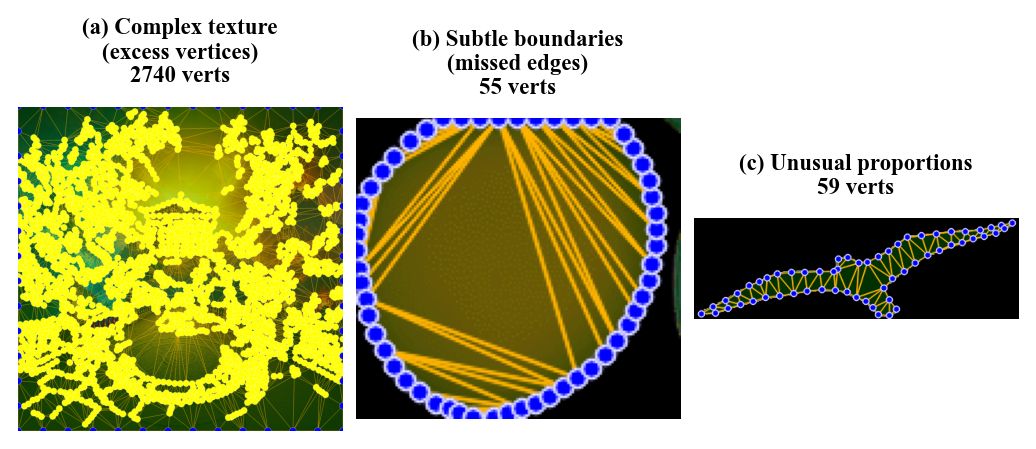}
    \caption{Examples of current limitations, shown as three mesh prediction results side by side with vertex counts. (a)~A highly textured sprite that produces excessive interior vertices due to many spurious Canny edges. (b)~A sprite with subtle visual boundaries (soft color gradients) where edges fall below the Canny threshold and are missed. (c)~A sprite with unusual proportions or an art style underrepresented in the training data, leading to imperfect segmentation or vertex placement.}
    \label{fig:limitations}
\end{figure}

\paragraph{3D Extension.}
While our work focuses on 2D meshes for skeletal animation, the principles---learned segmentation combined with contour-aware vertex placement---could be extended to 3D mesh generation from multi-view images or depth maps.
Figure~\ref{fig:limitations} illustrates representative examples of these failure modes.

\section{Conclusion}
\label{sec:conclusion}

We presented \method{}, a fully automatic pipeline for generating triangle meshes from 2D sprite images for skeletal animation.
Our approach combines neural network-based segmentation with classical computer vision algorithms for contour-aware vertex placement, bridging the gap between learned and algorithmic methods.

Through controlled experiments, we demonstrated that direct vertex position prediction via heatmap regression is not viable for this task, motivating our hybrid design.
Our segmentation network achieves an IoU of 0.87 on a novel dataset of over 100,000 sprite-mask pairs, and our contour-aware vertex placement algorithm produces meshes that faithfully follow the visual structure of input sprites.

\begin{figure}[H]
    \centering
    \includegraphics[width=\columnwidth,height=0.28\textheight,keepaspectratio]{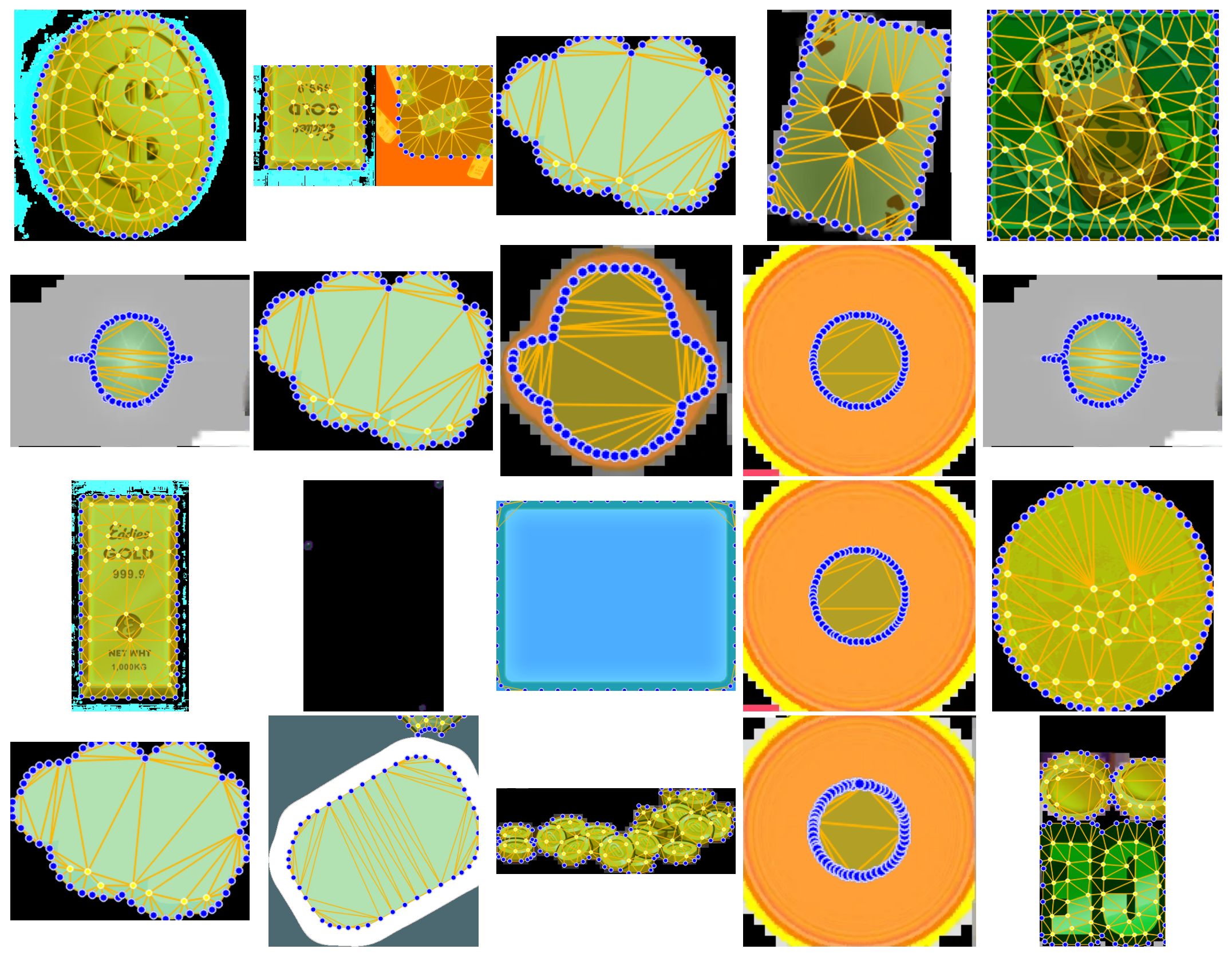}
    \caption{Gallery of automatically generated meshes arranged in a $4 \times 5$ grid. Each cell shows a single sprite with the predicted mesh overlaid as a wireframe. Our pipeline handles characters, objects, effects, and UI elements from diverse games without any manual intervention or parameter tuning.}
    \label{fig:gallery}
\end{figure}

The complete pipeline processes sprites in under 3 seconds, offering a speedup of several orders of magnitude over manual mesh creation, while producing results of comparable quality.
We release our code, trained model, and dataset to support further research in automated 2D animation asset production.
Figure~\ref{fig:gallery} presents a gallery of results across diverse sprites.

\bibliographystyle{plain}
\bibliography{references}

\end{document}